%% file: arxiv.tex
\documentclass[oneside,11pt]{article} 

\renewenvironment{abstract}
  {{\centering\large\bfseries Abstract\par}\vspace{0.7ex}%
    \bgroup
       \leftskip 20pt\rightskip 20pt\small\noindent\ignorespaces}%
  {\par\egroup\vskip 0.25ex}

{\par\egroup\vskip 0.25ex}

\usepackage[utf8]{inputenc} 
\usepackage[T1]{fontenc}  
\usepackage{microtype}

\usepackage{amsfonts,amssymb,amsmath,amsthm}
\usepackage{nicefrac}
\usepackage{xcolor}
\usepackage{url}
\usepackage{booktabs}
\usepackage{enumitem}

\usepackage{graphicx}
\usepackage{hyperref}

\numberwithin{equation}{section}
\theoremstyle{plain}
\newtheorem{theorem}{Theorem}[section]
\newtheorem{lemma}[theorem]{Lemma}
\newtheorem{proposition}[theorem]{Proposition}
\newtheorem{corollary}[theorem]{Corollary}

\theoremstyle{definition}
\newtheorem{remark}[theorem]{Remark}
\theoremstyle{definition}

\newcommand{\defeq}{:=}
\renewcommand{\(}{\left(}
\renewcommand{\)}{\right)}

\DeclareMathOperator*{\argmin}{arg\,min}

\newcommand{\st}{\operatorname{s.t.}}

\newcommand{\ba}{\boldsymbol{a}}

\newcommand{\bu}{\boldsymbol{u}}
\newcommand{\bv}{\boldsymbol{v}}

\newcommand{\bx}{\boldsymbol{x}}
\newcommand{\by}{\boldsymbol{y}}
\newcommand{\bz}{\boldsymbol{z}}

\newcommand{\bA}{\boldsymbol{A}}

\newcommand{\bI}{\boldsymbol{I}}

\newcommand{\bzero}{\boldsymbol{0}}

\newcommand{\xbar}{\bar{\bx}}

\newcommand{\xhat}{\hat{\bx}}

\newcommand{\calS}{\mathcal{S}}

\newcommand{\R}{\mathbb{R}}

\newcommand{\Rd}{\mathbb{R}^d}

\newcommand{\zeronorm}[1]{\left\lVert #1 \right\rVert_{0}}
\newcommand{\twonorm}[1]{\left\lVert #1 \right\rVert}
\newcommand{\onenorm}[1]{\left\lVert #1 \right\rVert_{1}}
\newcommand{\abs}[1]{\left\lvert #1 \right\rvert}

\newcommand{\infnorm}[1]{\left\lVert #1 \right\rVert_{\infty}}

\newcommand{\supp}[1]{\operatorname{supp}\(#1\)}

\newcommand{\trans}{^{\top}}
\newcommand{\inner}[2]{\left\langle #1, #2 \right\rangle}
\newcommand{\EXP}{\mathbb{E}}

\newcommand{\sign}[1]{\operatorname{sign}\(#1\)}

\newcommand{\const}{\mathrm{{C}}}

\newcommand{\Hk}[1]{\mathcal{H}_{k}\(#1\)}

\graphicspath{{fig/}}

\title{One-Bit Compressed Sensing via One-Shot Hard Thresholding}

\author{ 
Jie Shen\\
Stevens Institute of Technology\\
New Jersey, USA\\
\texttt{jie.shen@stevens.edu}
}

\begin{document}

\maketitle

\input{content}

\bibliography{../../../jshen_ref}
\bibliographystyle{plain}

\end{document}

%% file: content.tex
\begin{abstract}
This paper concerns the problem of 1-bit compressed sensing, where the goal is to estimate a sparse signal from a few of its binary measurements. We study a non-convex sparsity-constrained program and present a novel and concise analysis that moves away from the widely used notion of Gaussian width. We show that with high probability a simple algorithm is guaranteed to produce an accurate approximation to the normalized signal of interest under the $\ell_2$-metric. On top of that, we establish an ensemble of new results that address norm estimation, support recovery, and model misspecification. On the computational side, it is shown that the non-convex program can be solved via one-step hard thresholding which is dramatically efficient in terms of time complexity and memory footprint. On the statistical side, it is shown that our estimator enjoys a near-optimal error rate under standard conditions. The theoretical results are substantiated by numerical experiments.
\end{abstract}

\section{Introduction}\label{sec:intro}

The last two decades have witnessed a large demand of learning from high-dimensional data where the number of attributes is of the same order of, or even greater than the number of observations. Consider, for example, the Lou Gehrig's disease: there are millions of possible factors to evaluate but the scientists have a very restricted number of samples for research (each year 2 out of 100,000 individuals are affected by it). It thus turns out to be indispensable to investigate and to resolve two fundamental problems in this scenario: {\bfseries a)} when is it possible to learn a useful model from the small amount of data; and {\bfseries b)} how can we construct an accurate estimator with mild computational overhead. Answering these two questions has become a central theme in the field of high-dimensional statistics, and there have been a fruitful literature in a variety of applications such as linear regression \cite{chen1998atomic,tibshirani1996regression}, classification \cite{fan2008high}, principal component analysis \cite{ma2013sparse,yuan2013truncated}, precision matrix estimation \cite{cai2011constrained}, to name just a few.

As a special instance of linear regression, compressed sensing~(CS) has attracted increasing attention owing to its ease of implementation and the success in practical problems~\cite{donoho2006compressed}. To be more detailed, CS consists of two phases: data acquisition and signal recovery. Suppose that $\xbar \in \Rd$ is the signal of interest. During the first phase, the goal is to efficiently sample it and store the obtained measurements in a device. In light of the fact that such a device may not be computationally powerful, a common paradigm is to adopt linear measurements which mitigate hardware implementation and accelerate data collection. That is, we specify the number of measurements $n$ and the sensing vectors $\{ \ba_i \}_{i=1}^n \subset \Rd$, and record in the device the measurements $\{ y_i \}_{i=1}^n$ given by
\begin{equation}\label{eq:linear model}
y_i = \inner{\ba_i}{\xbar},\ \forall\ i = 1, \dots, n.
\end{equation}
For the purpose of efficient sampling, it is required that $n \ll d$, hence the name compressed sensing.
 
During the second phase, one has access to $\{ (\ba_i, y_i ) \}_{i=1}^n$ and the primary concern is to recover the underlying signal $\xbar$ on a possibly powerful machine. While \eqref{eq:linear model} is an underdetermined linear system that might have infinite number of feasible solutions, it has been well-understood that with a careful design of the sensing vectors, a broad range of algorithms are capable of producing accurate, or even exact reconstruction of $\xbar$ in polynomial time provided that it exhibits certain low-dimensional structure. See, for example, \cite{candes2005decoding,tropp2007signal,wainwright2009sharp,blumensath2009iterative,shen2018tight}.

Though elegant in theory and compelling in practice, \cite{boufounos2008bit} pointed out that it is not always realistic to collect the measurements as in~\eqref{eq:linear model} since they entail infinite-bit precision for the hardware. Alternatively, the measurements are often quantized into finite bits and in the extreme case, only the sign patterns are retained:
\begin{equation}\label{eq:1-bit linear}
y_i = \sign{ \inner{\ba_i}{\xbar} },\ \forall\ i = 1, \dots, n.
\end{equation}
The problem of recovering $\xbar$ from its binary measurements is referred to as 1-bit compressed sensing, and it bears the potential of savings of physical storage as long as accurate estimation in the 1-bit setting does not require significantly more measurements~\cite{jacques2013robust}. Note that as the sign function absorbs the magnitude of $\xbar$, we can only hope to reconstruct the direction $\xbar / \twonorm{\xbar}$ in general. In this light, the research lines can be roughly divided into three spaces: {\bfseries (a)}~sparse approximation, i.e. identifying the direction; {\bfseries (b)}~norm estimation, i.e. evaluating the norm with extra information; and {\bfseries (c)}~support recovery, i.e. determining the position of the non-zero entries of $\xbar$. 

\paragraph{Sparse Approximation.}
Perhaps surprisingly, it was provably shown in \cite{plan2013robust} that by seeking a global optimum of $\ell_1$-norm constrained programs, it is possible to reliably recover the underlying normalized signal with as many observations as in the standard model~\eqref{eq:linear model}. However, it comes up with a computational issue when optimizing the programs over large-scale data sets. From a technical perspective, since the programs are not strongly convex, only a sublinear convergence rate is guaranteed for projected gradient descent~\cite{nesterov2004introductory}. In other words, one has to run the algorithm for a considerable number of iterations in order to attain the optimum. Therefore, a large body of works were dedicated to investigating alternative programs that are endowed with closed-form solutions, which naturally rules out the trouble of slow convergence. For instance, \cite{zhang2014efficient} considered optimizing an $\ell_1$-norm penalized function which is exactly the Lagrangian function of the program in \cite{plan2013robust}. Interestingly, with the slight modification, it was shown that a simple element-wise thresholding on a certain vector gives even better error rate. Yet, for the sake of accurate recovery, one has to search for a proper regularization parameter which itself could be expensive. In particular, specifying a large value for it will result in trivial statistical guarantee whereas a small value will lead to a trivial solution (i.e. a zero vector).  \cite{huang2018nonconvex} illustrated that one can derive closed-form solutions for other sparsity-regularized programs such as smoothly clipped absolute deviation penalty \cite{fan2001variable}, minimax concave penalty \cite{zhang2010nearly}, and $\ell_0$-norm (that counts the number of non-zeros of a vector). Unfortunately, no statistical guarantee was established in that work.

\paragraph{Norm Estimation.}
While it seems implausible to evaluate the norm of $\xbar$ under the 1-bit setting, a number of recent works asserted that the premise quickly changes under extra conditions. For example, \cite{knudson2016bit,baraniuk2017exponential} concurrently showed that by manually adding Gaussian noise before quantization, one may estimate the magnitude by solving an augmented program that incorporates the noise information. This idea was further utilized to deal with heavy-tailed sensing vectors in \cite{dirksen2018robust}.

\paragraph{Support Recovery.}
Parallel to the line of estimating the direction of the signal, a plethora of works examined the problem of recovering the support set. In statistics, it is also known as variable selection or feature selection~\cite{tibshirani1996regression,wang2018provable}. Note that support recovery sheds light on sparse approximation in that once the support is identified, we can safely eliminate the irrelevant features and apply standard tools from regression theory (that has a rich literature). Unfortunately, it is typically more challenging to establish theoretical guarantee in this space even under the standard CS model~\cite{yuan2016exact,shen2017iteration,shen2017partial}. In the regime of 1-bit CS, \cite{gopi2013bit,acharya2017improved} presented a set of impressive results based on a new family of sensing vectors that move away from the standard Gaussian. The data acquisition is computationally more demanding though.

\subsection{Our Contributions}
We propose to study a sparsity-constrained non-convex program for the 1-bit CS problem. We first demonstrate that the global optimum can be computed by a simple hard thresholding operator (to be defined) which is computationally efficient. More importantly, the solution is proved to have a near-optimal approximation error rate to the direction of $\xbar$. We develop novel analysis showing that the approximation error is controlled by two factors: one is independent of the signal structure while the other is entirely determined by it. On top of that, we provide theoretical justification that our estimator recovers the support set of $\xbar$, can be tailored to offer an accurate estimation of the magnitude of $\xbar$, and is resilient to model misspecification. Namely, even when we feed an inappropriate parameter to the program, our estimate still exhibits favorable performance.

\subsection{Related Works}


The 1-bit CS problem is closely related to learning halfspaces~--~a central object of study in learning theory~\cite{valiant1984theory}. In compressed sensing, there is an additional structural assumption that the halfspace is embedded in a high-dimensional space, and at the heart of CS is leveraging such prior knowledge for improved sample complexity. In view of the connection to learning theory, a surge of recent works are devoted to understanding their interplay. For example, active learning is a long-established research field in learning theory, which attempts to mitigate the human labor for data annotation by actively querying the labels~\cite{awasthi2017power,yan2017revisiting,zhang2018efficient,zhang2020efficient,shen2020attribute}. Similar ideas are also explored in CS, known as adaptive sensing~\cite{castro2013fundamental,baraniuk2017exponential}.

It is also worth mentioning that one can expand the observation model~\eqref{eq:1-bit linear} to 1-bit matrix completion that has found successful applications in social networks. For example, in recommender systems we can construct an incomplete data matrix where each observed entry represents the preference (like or dislike) of a user to an item~\cite{davenport2014bit,wang2017online}. The goal is to predict the missing entries and to recommend items that are potentially interesting for the customers. The statistical guarantee for this problem has been established in a series of appealing works~\cite{bhaskar2016probabilistic,shen2019robust}.

\paragraph{Notation.}
We use bold lowercase letters such as $\bv$ to denote a column vector. Its $i$th element is denoted by $v_i$. There are three norms that will be involved for a vector $\bv \in \Rd$: the $\ell_2$-norm $\twonorm{\bv} \defeq \sqrt{ v_1^2 + \dots + v_d^2 }$, the $\ell_1$-norm $\onenorm{\bv} \defeq \abs{v_1} + \dots + \abs{v_d}$, and the $\ell_{\infty}$-norm $\infnorm{\bv} \defeq \max_{1 \leq i \leq d} \abs{v_i}$. We write the number of non-zero elements in $\bv$ as $\zeronorm{\bv}$, and with a slight abuse of terminology we will call it $\ell_0$-norm\footnote{$\zeronorm{\bv}$ is not a norm as it is not absolutely homogeneous.}.

For a finite set $\calS$, its cardinality is denoted by $\abs{\calS}$. The index set of the non-zero elements of a vector $\bv \in \Rd$ is called the support set, and is denoted by $\supp{\bv}$. We write the index set of the top $k$ elements (in magnitude) as $\supp{\bv, k}$ with ties being broken lexicographically. 

We reserve $\xbar$ for the $s$-sparse target signal that we hope to estimate. We also reserve the upright capital letter $\const$ and its subscript variants such as $\const_0$ and $\const_1$ for absolute constants, whose values may change from appearance to appearance. For two scalars $a$ and $b$, we write $a = O(b)$ if $a \leq \const \cdot b$; and we write $a = \Omega(b)$ if $a \geq \const \cdot b$.

\paragraph{Overview.}
In Section~\ref{sec:prob}, we elaborate the problem setup of 1-bit compressed sensing. The primary theoretical results and the comparison to the prior works are developed in Section~\ref{sec:main}, and Section~\ref{sec:ext} presents useful extensions of our main results. In Section~\ref{sec:exp} we provide empirical evidence to support our analysis. Section~\ref{sec:con} concludes the paper and all the proof details are deferred to the Appendix (see the supplementary material).

\section{Preliminaries}\label{sec:prob}

Let $\xbar \in \Rd$ be the $s$-sparse signal of interest. For now, we presume without loss of generality that it has unit $\ell_2$-norm. While our concentration is on the 1-bit CS problem~\eqref{eq:1-bit linear}, we begin our discussion with a more general observation model: for each measurement $y_i \in \{-1, 1\}$, instead of treating it as being generated from a deterministic mapping through the sign function, we consider
\begin{equation}\label{eq:model yi}
\EXP[y_i \mid \ba_i ]  = \theta_i\( \inner{\ba_i}{\xbar} \),\ \forall\ i=1, \dots, n,
\end{equation}
where the mapping functions $\theta_i(\cdot) \in [-1, +1]$. Note that under the observation model~\eqref{eq:1-bit linear}, $\theta_i(\cdot)$ is exactly the sign function. It is worth mentioning that our model~\eqref{eq:model yi} is more general than the one considered in prior works~\cite{plan2013robust,knudson2016bit,plan2017high}, since we allow different mapping function for different sample, i.e. $\theta_i$ can be mutually distinct.

\subsection{The Non-Convex Estimator}

In order to recover $\xbar$, we consider the estimate $\xhat$ that is a global optimum of the following program:
\begin{equation}\label{eq:main prog}
\max_{\bx \in \Rd}\ \by\trans \bA \bx,\quad \st\ \zeronorm{\bx} \leq k, \ \twonorm{\bx} = 1,
\end{equation}
where $\by = (y_1, \dots, y_n)\trans$ and $\bA = (\ba_1, \dots, \ba_n)\trans$. The two constraints accommodate our prior knowledge on the signal $\xbar$, and the objective function seeks for a maximum correlation between the input and output of $\theta_i(\cdot)$. We recall that $\xbar$ is $s$-sparse. The integer $k > 0$ serves as the only parameter in our estimator, and we need to tune it in practice. Ideally, $k$ plays as an upper bound on the true sparsity $s$, which indicates that our estimator is unbiased, i.e. the true signal is contained in the feasible set. Once the condition is violated, we say the model is misspecified which needs a particular treatment (see Section~\ref{subsec:misspec}).

The optimal solution of the program~\eqref{eq:main prog} can be obtained by simple algebraic calculation. Define the hard thresholding operator as follows:
\begin{equation}
\Hk{\bz} \defeq \argmin_{\bu \in \Rd, \zeronorm{\bu} \leq k} \twonorm{\bu - \bz}.
\end{equation}
The following computational result is due to \cite{bahmani2013robust}. We present the proof in Appendix~\ref{sec:app:proof of xhat} for completeness.
\begin{proposition}\label{prop:xhat solution}
The global optimum $\xhat$ of the program~\eqref{eq:main prog} is given by
\begin{equation*}
\xhat = \frac{\Hk{ \bv } }{ \twonorm{  \Hk{ \bv } } },\ \text{where}\ \bv \defeq \bA\trans \by .
\end{equation*}
\end{proposition}
Observe that the time complexity of calculating $\bA\trans\by$ is $O(dn)$, and applying the hard thresholding costs $O( k \log d)$ since the $k$-sparse vector $\Hk{\bz}$ can be efficiently computed by first sorting the elements of $\bz$ in their magnitude, and then retaining the top $k$ of them. Therefore, the total running time is $O(dn)$ given that $n$ is always greater than $k \log d$ in compressed sensing. 

\section{Main Results}\label{sec:main}

We now move on to present the statistical estimation error under standard conditions~\cite{plan2013robust}. We assume that each observation $y_i$ depends on the measurement vector $\ba_i$ only through $\inner{\ba_i}{\xbar}$. Namely, 
\begin{enumerate}[label=$(A\arabic*)$, start=1, leftmargin=*]
\item given the inner product $\inner{\ba_i}{\xbar}$, $y_i$ and $\ba_i$ are conditionally independent.\label{as:indep}
\end{enumerate}

We will focus on the standard Gaussian design of the sensing vectors. That is,
\begin{enumerate}[label=$(A\arabic*)$, start=2, leftmargin=*]
\item $\ba_i \sim N\(\bzero, \bI_{d\times d}\)$ for all $i = 1, \dots, n$, and they are mutually independent. \label{as:ai}
\end{enumerate}
The above sensing scheme offers a clean picture of theoretical understanding. Furthermore, in the context of compressed sensing, we indeed have the control of selecting the sensing vectors. Note that it is possible to relax the assumption to correlated Gaussian design~\cite{plan2013robust}, or even non-Gaussian measurements~\cite{ai2014one,dirksen2018robust}. 

In order to estimate the $s$-sparse signal $\xbar$ from its nonlinear measurements, we have to confine ourselves to a family of mapping functions. As a matter of fact, if all functions $\theta_i$ output zero almost surely, then no algorithm is able to recover the underlying signal. Mathematically, the minimum requirement on the functions is that their input and output are correlated. Let
\begin{equation}\label{eq:lambda}
\lambda_i \defeq \EXP_{g \sim N(0, 1)}\big[ g \cdot \theta_i(g) \big],\ \forall\ i=1, \dots, n.
\end{equation}
In light of~\ref{as:indep} and~\ref{as:ai}, $\lambda_i$ essentially characterizes the correlation of interest~\cite{plan2013robust}. While we do not impose individual condition on $\lambda_i$, we need to assume
\begin{enumerate}[label=$(A\arabic*)$, start=3, leftmargin=*]
\item the average correlation $\lambda \defeq \frac{1}{n} \sum_{i=1}^{n}\lambda_i > 0$. \label{as:lambda}
\end{enumerate}
Again, since we have the freedom to design $\theta_i(\cdot)$, we can always replace all $\theta_i(\cdot)$ with $-\theta_i(\cdot)$ if we find $\lambda < 0$. We remark that a very recent work~\cite{thrampoulidis2019lift} studies the interesting case $\lambda = 0$ under extra assumptions.

We are now in the position to present performance guarantee of our estimate $\xhat$. We show that as soon as we collect $n = O(k \log d)$ measurements, it is possible to accurately approximate the direction of $\xbar$ with high probability from its non-linear measurements.

\begin{theorem}\label{thm:main}
Assume \ref{as:indep}, \ref{as:ai} and \ref{as:lambda}. Further assume $\twonorm{\xbar} = 1$, $\zeronorm{\xbar} = s$, and $k \geq s$. With probability at least $1 - d^{-10}$, we have
\begin{equation*}
\twonorm{\xhat - {\xbar}} \leq \frac{\const}{\lambda} \sqrt{\frac{k \log d}{n}}.
\end{equation*}
\end{theorem}
\begin{remark}
While our assumption on $\lambda$ is very mild, it is important to note that in order to obtain near-optimal sample complexity, $\lambda$ must act as a universal constant, which translates into an implicit requirement that most of the $\lambda_i$'s are positive constants. Otherwise, Theorem~\ref{thm:main} may offer trivial guarantee. Consider, for example, $\lambda_1 = \lambda_2 = \dots \lambda_{n-1} = 0$ and $\lambda_n = 1$. It corresponds to $\lambda = 1/n$ for which we have a trivial upper bound $O(\sqrt{n k \log d})$ on the estimation error. This is not surprising since many zero correlations indicate that the sampling power is wasted on the associated samples (and thus recovery is impossible).
\end{remark}

Fortunately, we can show that among many prevalent statistical models, the parameter $\lambda$ acts as a universal non-zero constant, which indicates that the sample complexity of our estimator is near-optimal in light of the fact that the information-theoretic lower bound for infinite-bit CS is $\Omega(k \log (d/k))$~\cite{raskutti2011minimax}.

\subsection{Noiseless 1-Bit CS}
Consider the problem~\eqref{eq:1-bit linear} where $\theta_i(g) = \sign{g}$. With some calculation,
\begin{equation*}
\lambda_i = \EXP_{g \sim N(0, 1)}[ g \sign{g}] = \EXP[ \abs{g}] = \sqrt{2/\pi}.
\end{equation*}
Thus, we obtain the following corollary regarding the sample complexity of our estimator.
\begin{corollary}
Assume the same conditions as in Theorem~\ref{thm:main}. Under the model~\eqref{eq:1-bit linear},
\begin{equation*}
\Pr\(\twonorm{\xhat - \xbar} \leq \epsilon \) \geq 1 - d^{-10}
\end{equation*}
for any $\epsilon \in (0, 1)$ provided that $n = O\({\epsilon^{-2}} k \log d\)$.
\end{corollary}

\subsection{Noisy 1-Bit CS}
In real-world applications, the observations are not only highly quantized, but are also grossly corrupted. In fact, owing to possible systematic errors or human mistakes, the sign may be flipped with some probability. The noisy model can thus be formulated as
\begin{equation}\label{eq:1-bit linear noisy}
y_i = \xi_i \sign{\inner{\ba_i}{\xbar}},\quad \forall\ i = 1,\ \dots,\ n,
\end{equation}
where $\xi_i$ is independent of $\ba_i$ and
\begin{equation*}
\Pr(\xi_i = 1) = 1 - p_i,\quad \Pr(\xi_i = -1) = p_i
\end{equation*}
for some $p_i \in [0, 0.5)$. This gives a new mapping function and a new correlation parameter as follows:
\begin{align*}
& \theta_i(g) = \sign{g} \cdot \EXP[\xi_i] = (1 - 2p_i) \sign{g}, \\
& \lambda_i = \sqrt{2/\pi} (1 - 2p_i).
\end{align*}

\begin{corollary}
Assume the same conditions as in Theorem~\ref{thm:main}. Under the model~\eqref{eq:1-bit linear noisy}, for any $\epsilon \in (0, 1)$
\begin{equation*}
\Pr\(\twonorm{\xhat - \xbar} \leq \epsilon \) \geq 1 - d^{-10}
\end{equation*}
provided that $n = O\({\((1-2p)\epsilon\)^{-2}} k \log d\)$ where $p = \frac{1}{n}\sum_{i=1}^{n}p_i$.
\end{corollary}


\subsection{Proof Sketch of Theorem~\ref{thm:main}}
Here we sketch the proof and highlight our novelty for the analysis. We defer all technical details to Appendix~\ref{sec:app:proof}. Note that $\xbar$ is a feasible solution to \eqref{eq:main prog}. Owing to the optimality of $\xhat$, it holds that
\begin{equation*}
\by\trans \bA \xhat \geq \by\trans \bA \xbar.
\end{equation*}
With some re-arrangement, we have
\begin{equation*}
 \inner{- \lambda \xbar}{\xhat - \xbar} \leq \inner{\frac{1}{n} \bA\trans \by - \lambda \xbar}{\xhat - \xbar}.
\end{equation*}
Since $\twonorm{\xbar} = \twonorm{\xhat} = 1$, it follows that $\inner{-\lambda \xbar}{\xhat - \xbar} = \frac{\lambda}{2} \twonorm{\xhat - \xbar}^2$. Thus, we obtain
\begin{align}\label{eq:tmp99}
\frac{\lambda}{2} \twonorm{\xhat - \xbar}^2 \leq&\ \inner{\frac{1}{n} \bA\trans \by - \lambda \xbar}{\xhat - \xbar} \notag\\
\leq&\ \infnorm{\frac{1}{n} \bA\trans \by - \lambda \xbar } \cdot \onenorm{ \xhat - \xbar}\notag\\
\leq&\ \infnorm{\frac{1}{n} \bA\trans \by - \lambda \xbar } \cdot \sqrt{2k} \twonorm{\xhat - \xbar},
\end{align}
where the second inequality follows from H\"older's inequality. For the third inequality, it follows from the facts that $\onenorm{\bv} \leq \sqrt{\zeronorm{\bv}} \cdot \twonorm{\bv}$ for all $\bv \in \Rd$ and that $\zeronorm{\xhat - \xbar} \leq 2k$. Dividing both sides by $\twonorm{\xhat - \xbar}$ gives
\begin{equation}\label{eq:bound}
\twonorm{\xhat - \xbar} \leq { \frac{\sqrt{2k}}{\lambda} \infnorm{\frac{1}{n} \bA\trans \by - \lambda \xbar } }.
\end{equation}
It remains to upper bound the infinity norm.
\begin{lemma}[\cite{zhang2014efficient}]\label{lem:mean}
Assume \ref{as:indep} and \ref{as:ai}. Then for all $i = 1, \dots, n$, it holds that $\EXP[\ba_i y_i] = \lambda_i \xbar$.
\end{lemma}
In particular, Lemma~\ref{lem:mean} implies $\frac{1}{n} \bA\trans \by - \lambda \xbar = \frac{1}{n} \sum_{i=1}^{n} ( \ba_i y_i - \EXP[\ba_i y_i])$. Hence, we can apply Hoeffding's inequality to show that the sum of independent random variables concentrates around its mean with high probability. This is formally stated below.
\begin{lemma}\label{lem:inf norm}
Consider the observation model~\eqref{eq:model yi}. Assume \ref{as:indep}, \ref{as:ai} and \ref{as:lambda}. With probability at least $1 - d^{-10}$ (over the random draw of $\bA$) it holds that
\begin{equation*}
\infnorm{\frac{1}{n} \bA\trans \by - \lambda \xbar } \leq \const \sqrt{\frac{\log d}{n}}
\end{equation*}
for some absolute constant $\const > 0$.
\end{lemma}
Applying the inequality of the above lemma, we obtain Theorem~\ref{thm:main}.\hfill$\qed$

\begin{remark}[Decomposition of Estimation Error]\label{rmk:decomp}
As can be seen from the proof, if $\xbar$ belongs to a generic constraint set $\mathcal{K}$, then we can tailor our analysis as follows. First, we will solve
\begin{equation*}
\max_{\bx} \ \by\trans\bA\bx, \quad \st\ \bx \in \mathcal{K}.
\end{equation*}
We present the closed-form solution for certain $\mathcal{K}$ (a solution for general $\mathcal{K}$ is hard to derive). Let $\bv = \bA\trans \by$.
\begin{itemize}
\item $\mathcal{K} = \{\bx \in \Rd: \zeronorm{\bx} \leq k, \twonorm{\bx} = 1, \bx \geq \bzero\}$. Let $S = \{i: v_i > 0\}$ and $m = \abs{S}$. If $m > k$, redefine $S$ as the index set of the $k$ largest elements of $\bv$. The global optimum $\xhat$ for such $\mathcal{K}$ is given as follows: if $m = 0$, then $\xhat$ is the $i$th standard basis vector where $i$ is the index of the largest entry of $\bv$; otherwise $\xhat = \bv_S / \twonorm{\bv_S}$ where $\bv_S$ is obtained by setting all elements of $\bv$ outside $S$ to zero.

\item $\mathcal{K} = \{\bx: \zeronorm{\bx} \leq k, \bx \in \{-1, 0, 1\}^d \}$. Let $S = \supp{\bv, k}$. Then $\xhat = \sign{\bv_S}$.
\end{itemize}
Now suppose that we are able to obtain a global optimum $\xbar$. Then following the same reasoning, \eqref{eq:tmp99} becomes
\begin{equation}\label{eq:tmp98}
\frac{\lambda}{2} \twonorm{\xhat - \xbar}^2 \leq \infnorm{\frac{1}{n} \bA\trans \by - \lambda \xbar } \cdot \rho_{\mathcal{K}}^{} \twonorm{\xhat - \xbar},
\end{equation}
where
\begin{equation}
\rho_{\mathcal{K}}^{} \defeq \sup_{\bz \in \mathcal{K} - \mathcal{K}} \ \frac{\onenorm{\bz}}{\twonorm{\bz}}
\end{equation}
is the restricted induced norm which is completely characterized by the signal structure $\mathcal{K}$. While $\rho_{\mathcal{K}}^{} \leq \sqrt{\zeronorm{z}}$ for general $\mathcal{K}$, it is possible to show an improved bound for specific $\mathcal{K}$. For instance, if $\mathcal{K}$ is the set of $s$-sparse $\alpha$-strongly-decaying signals~\cite{davenport2010analysis} for some $\alpha \geq 2$, then $\rho_{\mathcal{K}}^{}$ acts as an absolute constant, which implies that the sample complexity in Theorem~\ref{thm:main} can be improved to $O(\log d)$. On the other hand, Lemma~\ref{lem:inf norm} tells us that the infinity norm in \eqref{eq:tmp98} is oblivious of $\mathcal{K}$ (but depends on the distribution of $\bA$ and observation model). Therefore, if one is interested in a sensing matrix $\bA$ with heavy-tailed distributions, it suffices to derive a new bound as what we did in Lemma~\ref{lem:inf norm}.
\end{remark}

\subsection{Comparison to Prior Works}

We first compare with two of the most important works on 1-bit CS~\cite{plan2013robust,plan2017high}. Both of them proposed to recover a signal with a generic structure $\mathcal{K}$. Regarding theoretical guarantee, \cite[Theorem~1.1]{plan2013robust} implies that to obtain $\twonorm{\xhat - \xbar} \leq \epsilon$, the sample size $n = O\(\epsilon^{-4} k \log(d/k) \)$ which is worse than what we derived in Theorem~\ref{thm:main} in terms of the dependence on $\epsilon$. Though \cite[Theorem~2.1]{plan2017high} improved the sample complexity to $n = O\(\epsilon^{-2} k \log(d/k) \)$, their proof is technically involved (e.g. they applied high-dimensional geometric arguments) even specifying $\mathcal{K}$ as the sparsity constraint. In contrast, we depart from their theoretical analysis and reach the same guarantee with fundamental facts in probability theory. Our analysis (Remark~\ref{rmk:decomp}) shows the decomposition of estimation error which is useful to develop new results for different sensing schemes and signal structures. 

In the seminal work of~\cite{jacques2013robust}, a lower bound on the statistical error of sparse approximation was established, and a non-convex program was proposed to achieve the lower bound up to some logarithmic factor. Notably, their guarantee is uniform, meaning that a single draw of the sensing vectors ensures recovery of all sparse signals, whereas our result is specified to a particular signal. Their sample complexity is $n = O(\epsilon^{-1} k \log d)$, which has a better dependence on $\epsilon$ than this work. However, due to the non-convex nature, it is not clear how to solve their program in polynomial time. As a matter of fact, an iterative algorithm was devised with compelling performance in practical problems, but it lacks theoretical backend on the convergence behavior.

It is worth noting that \cite[Theorem~4.3]{zhu2015towards} claimed a sample complexity bound of $O(s)$ for certain type of $\mathcal{K}$, which is a stronger result than what we obtained in the paper. However, this may not be the true sample complexity since in order to fulfill their conditions (i.e. $\xbar_{\min}$ is sufficiently large), one needs $O(s \log d)$ samples. For the recovery of general sparse signals, they showed a sample complexity bound similar to ours through different non-convex estimators. In this case, our primary improvement falls into the modular analysis and computational efficiency.


\section{Extensions}\label{sec:ext}

Theorem~\ref{thm:main} offers near-optimal guarantee on recovering the direction of the signal $\xbar$. In this section, we discuss when we can recover its support set, and even its magnitude under extra conditions. In this section, we focus on the elementary case that $\xbar$ is $s$-sparse.

\subsection{Support Recovery}

We first describe when our estimator reliably detects the support of $\xbar$. We would like point out that in some applications such as medical diagnosis, it is of crucial importance to discover the determinants of a disease (i.e. support recovery). Intuitively, a factor can be identified only when it has ``sufficient'' impact on the disease. This notion of significance is characterized by the following mathematical quantity:
\begin{equation}\label{eq:xmin}
\xbar_{\min} \defeq \min_{i \in \supp{\xbar} }  \abs{\bar{x}_i}.
\end{equation}
Throughout the paper, we consider $\xbar_{\min} \neq 0$, i.e. the signal of interest is non-zero. We utilize a well-known fact to derive the guarantee of support set of $\xbar$.
\begin{lemma}\label{lem:supp recovery}
For a given signal $\xbar$, if
\begin{equation*}
\twonorm{\bx - \xbar} < \xbar_{\min},
\end{equation*}
then it holds that $\supp{\xbar} \subset \supp{\bx}$.
\end{lemma}
The lemma can be proved by algebra (see Appendix~\ref{sec:app:proof of supp}), and has been widely used in the literature~\cite{wainwright2009sharp}. In allusion to Theorem~\ref{thm:main} (that is, let the right-hand side therein be less than $\xbar_{\min}$), we show that the support set of $\xbar$ is contained in that of our estimate.
\begin{theorem}\label{thm:supp}
Assume the same conditions as in Theorem~\ref{thm:main}. Then $\supp{\xbar} \subset \supp{\xhat}$ provided that
\begin{equation*}
\xbar_{\min} > \frac{\const}{\lambda} \sqrt{\frac{k \log d}{n}}.
\end{equation*}
In particular, if we know exactly the sparsity of $\xbar$, we have $\supp{\xbar} = \supp{\xhat}$.
\end{theorem}
\begin{remark}
In the special case that $\xbar$ is a binary vector, the above theorem indicates exact signal recovery under near-optimal sample complexity.
\end{remark}
\begin{remark}
The proof of Lemma~\ref{lem:supp recovery} essentially suggests that the minimum condition for support recovery is $\infnorm{\xhat - \xbar} < \xbar_{\min}$. Yet we conjecture that even with such condition Theorem~\ref{thm:supp} may not be significantly improved. Suppose that $N$ samples suffice for support recovery of $\xbar$. Consider a two-step scheme of sparse approximation: 1) recover support; 2) linear regression restricted on the obtained support set. Since the second step needs $k$ samples, the total sample size is $N+k$. As lower bound of sparse approximation is $k \log(d/k)$ we must have $N\geq k \log(d/k)$.
\end{remark}

\subsection{Norm Estimation}\label{subsec:norm}
In this section we consider $\twonorm{\xbar} \leq R$ where $R$ is known, and we hope to estimate the norm of $\xbar$. In general, this is impossible in that the sign function will absorb the magnitude information. Thus, we shall make a further assumption for the data acquisition procedure:
\begin{equation}\label{eq:obs norm}
y_i = \sign{\inner{\ba_i}{\xbar} + b_i},\ \forall\ i=1, \dots, n,
\end{equation}
where $b_i$ are manually added noise which is known to us. The above observation model is equivalent to
\begin{equation*}
y_i = \sign{\inner{\ba_i'}{\xbar'}},\ \forall\ i=1, \dots, n,
\end{equation*}
where
\begin{equation*}
\ba_i' = \begin{pmatrix}
\ba_i\\
b_i / R
\end{pmatrix},\quad
\xbar' = \frac{1}{\sqrt{\twonorm{\xbar}^2 + R^2}} \begin{pmatrix}
\xbar\\
R
\end{pmatrix}.
\end{equation*}
Note that the norm of $\xbar$ has been encoded into $\xbar'$ and $y_i$ bears the information of $b_i$, which together paves the way for norm estimation. Also, all the $\ba_i'$ are i.i.d. standard Gaussian random vectors provided that
\begin{enumerate}[label=$(A\arabic*)$, start=4, leftmargin=*]
\item $b_i \sim N\(0, R^2\)$ for all $i = 1, \dots, n$, they are mutually independent, and are independent from all $\ba_i$. \label{as:ai'}
\end{enumerate}
Since $\xbar'$ is an $(s+1)$-sparse signal, and has unit $\ell_2$-norm, the estimation of $\xbar'$ from \eqref{eq:obs norm} can be reduced to that of $\xbar$ from \eqref{eq:1-bit linear} through the following augmented program:
\begin{equation}\label{eq:aug prob}
\max_{{\bx'} \in \mathbb{R}^{d+1}}\ \by\trans {\bA'} {\bx'}, \quad \st\ \zeronorm{{\bx'}} \leq k,\ \twonorm{\bx'} = 1,
\end{equation}
where $\bA' = (\ba_1', \dots, \ba_n')\trans$ and $k \geq s+1$. Proposition~\ref{prop:xhat solution} immediately gives the global optimum of~\eqref{eq:aug prob}:
\begin{equation}\label{eq:xhat norm}
\xhat' = \Hk{(\bA')\trans \by} / \twonorm{\Hk{(\bA')\trans \by}}.
\end{equation}
With the optimal solution, we are able to evaluate the magnitude of $\xbar$, as presented in the following theorem.
\begin{theorem}\label{thm:norm}
Consider the statistical model~\eqref{eq:obs norm}. Assume \ref{as:indep}, \ref{as:ai}, \ref{as:lambda} and \ref{as:ai'}. Further assume that $\twonorm{\xbar} \leq R$, $\zeronorm{\xbar} = s$, and $k \geq s+1$. Write $\xhat'$ as
\begin{align*}
\xhat' = \frac{(\bx_0;\ t_0)}{\sqrt{ \twonorm{\bx_0}^2 + t_0^2 }}.
\end{align*}
With probability at least $1- d^{-10}$ over the random draw of $\bA$, we either have
\begin{equation*}
\twonorm{ \frac{R}{\twonorm{\bx_0}} \bx_0 - \xbar } \leq \frac{\const \cdot R}{\lambda} \sqrt{\frac{k\log d}{n}}
\end{equation*}
in the case $t_0 = 0$ (thus $\twonorm{\bx_0} \neq 0$); or have
\begin{equation*}
\twonorm{\frac{R}{t_0} \bx_0 - {\xbar}} \leq \frac{\const \cdot R}{\lambda} \sqrt{\frac{k\log d}{n}}
\end{equation*}
in the case $t_0 \neq 0$.
\end{theorem}
\begin{remark}
Interestingly, our theorem implies that when $t_0 = 0$ (i.e. the manually added noise does not play a role in estimation), $\frac{R}{\twonorm{\bx_0}} \bx_0$ is a good approximation to $\xbar$. In other words, the norm of $\xbar$ is extremely close (or even equal) to $R$. This result can be interpreted from another perspective: once we know the norm of $\xbar$ in advance, say $\twonorm{\xbar} = R$, it is easy to see that $\xbar / R$ is feasible to program~\eqref{eq:main prog}, and Theorem~\ref{thm:main} already implies
\begin{equation*}
\twonorm{\xhat - \frac{\xbar}{R}} \leq \frac{\const}{\lambda} \sqrt{\frac{k \log d}{n}},
\end{equation*}
which is precisely the first inequality in Theorem~\ref{thm:norm} by noticing $\bx_0 = \xhat$ when $t_0 = 0$. Namely, there is no need to consider model~\eqref{eq:obs norm}.

The analysis for $t_0 \neq 0$ follows from Theorem~4 in \cite{knudson2016bit}, which showed that as soon as we have a good approximation to the direction of $\xbar$, it is possible to estimate the magnitude if the conditions in Theorem~\ref{thm:norm} are fulfilled. However, the scenario $t_0 = 0$ was not addressed therein, and we make efforts to draw a formal analysis. See Appendix~\ref{sec:app:proof of norm} for a full proof.
\end{remark}

\subsection{Model Misspecification}\label{subsec:misspec}

In our previous discussion, we always assume that the sparsity $k$ is equal to, or greater than the true sparsity $s$, i.e. the true signal is contained in the feasible set. However, sometimes we may choose $k < s$ in that we are not aware of $s$. As a result, recovery of $\xbar$ is impossible but we can still hope to approximate its $k$ largest (i.e. most important) components to a high precision. 

We now elaborate the new results under the misspecified model. One notable fact is that even when $k < s$, the normalized sparse vector $\Hk{\xbar}$ is feasible to the non-convex program~\eqref{eq:main prog}. Therefore, for sparse approximation we may apply the same induction and obtain the following.
\begin{theorem}\label{thm:sparse approx misspec}
Assume the same conditions as in Theorem~\ref{thm:main} but $k < s$. With probability at least $1 - d^{-10}$,
\begin{equation*}
\twonorm{\xhat - \bz} \leq \frac{\const}{\lambda} \sqrt{\frac{k \log d}{n}} + \sqrt{2k} \infnorm{ \bz - \xbar},
\end{equation*}
where $\bz \defeq  \frac{\Hk{\xbar}}{\twonorm{\Hk{\xbar}}}$.
\end{theorem}
The proof can be found in Appendix~\ref{sec:app:proof of approx}. The second term on the right-hand side is the price we pay for model misspecification, and it vanishes as soon as $k \geq s$. It is worth mentioning that it depends exclusively on the nature of the signal $\xbar$ rather than on the data acquisition procedure. If the $s$-sparse signal $\xbar$ has a light tail, i.e. the first $k$ components dominate the magnitude, then $\xhat$ behaves as a good estimate. To see this, let us write $\xbar = (\bar{x}_1, \dots, \bar{x}_s, 0, \dots, 0)$ in descending order (according to the magnitude of the elements). In this way $\Hk{\xbar} = (\bar{x}_1, \dots, \bar{x}_k, 0, \dots, 0)$. Let us denote $\alpha = \twonorm{\Hk{\xbar}}$ for now. It follows that
\begin{equation}\label{eq:inf exp}
\infnorm{  \bz - \xbar } = \max\bigg\{ \(\frac{1}{\alpha} - 1\) \abs{\bar{x}_1},\ \abs{\bar{x}_{k+1}} \bigg\}.
\end{equation}
If the $k$ leading components dominate the remaining, then $\alpha \approx 1$ and $\abs{\bar{x}_{k+1}} \approx 0$ hold simultaneously. As a consequence, $\infnorm{\bz - \xbar}$ is close to zero, under which Theorem~\ref{thm:sparse approx misspec} implies that we can accurately identify the principal direction of $\xbar$ given sufficient measurements.

Now we move on to discuss the support recovery of the top $k$ elements. In light of Lemma~\ref{lem:supp recovery}, the support of $\xhat$ is consistent with that of $\Hk{\xbar}$ as soon as the sample size $n = O\( (\lambda \abs{\bar{x}_k})^{-2} k \log d \)$ and
\begin{equation*}
\sqrt{2k}\infnorm{  \bz - \xbar }  < \abs{\bar{x}_k}.
\end{equation*}
Consider in~\eqref{eq:inf exp} that the infinite norm is given by $(1/\alpha - 1) \abs{\bar{x}_1}$. It follows that a sufficient condition for support recovery is
\begin{equation*}
\abs{\bar{x}_{k+1}} \leq \( \frac{1}{\alpha} - 1\) \abs{\bar{x}_1} < \frac{\abs{\bar{x}_k} }{\sqrt{2k}}.
\end{equation*}
The first inequality, which upper bounds $\abs{\bar{x}_{k+1}}$, indicates that for our purpose, the elements outside of the support of interest cannot be too large. The second inequality, which lower bounds $\abs{\bar{x}_k}$, tells that those inside of the support need to have sufficient magnitude.

Lastly, as we discussed in Section~\ref{subsec:norm}, a good approximation to the direction of $\Hk{\xbar}$ implies a good estimation of the norm. We remark that this observation holds for a misspecified model as well.

\section{Experiments}\label{sec:exp}
This section is dedicated to examining the statistical error rate and robustness of the our estimator. We focus on the noiseless model~\eqref{eq:1-bit linear}, and will compare with the Lasso estimator~\cite{plan2017high} which showed state-of-the-art performance.

\paragraph{Settings.}
We implement our algorithm and the one of \cite{plan2017high} in Matlab 2018, and perform all the experiments on a single server which has two 3.2 GHz Intel Xeon processors, each of which has 8 cores. The sensing vectors are chosen as i.i.d. standard Gaussian. For the $s$-sparse signal $\xbar$, we first randomly choose the support set in a uniform manner, and draw each non-zero element from an i.i.d. uniform distribution over the interval $[-1000, 1000]$. For each experiment to be presented, we generate 100 i.i.d. copies of the true signal $\xbar$ and report the averaged performance. If not specified, we always set $k = s$.

\begin{figure}[h]
\vspace{-0.05in}
\centering
\includegraphics[width=0.4\linewidth]{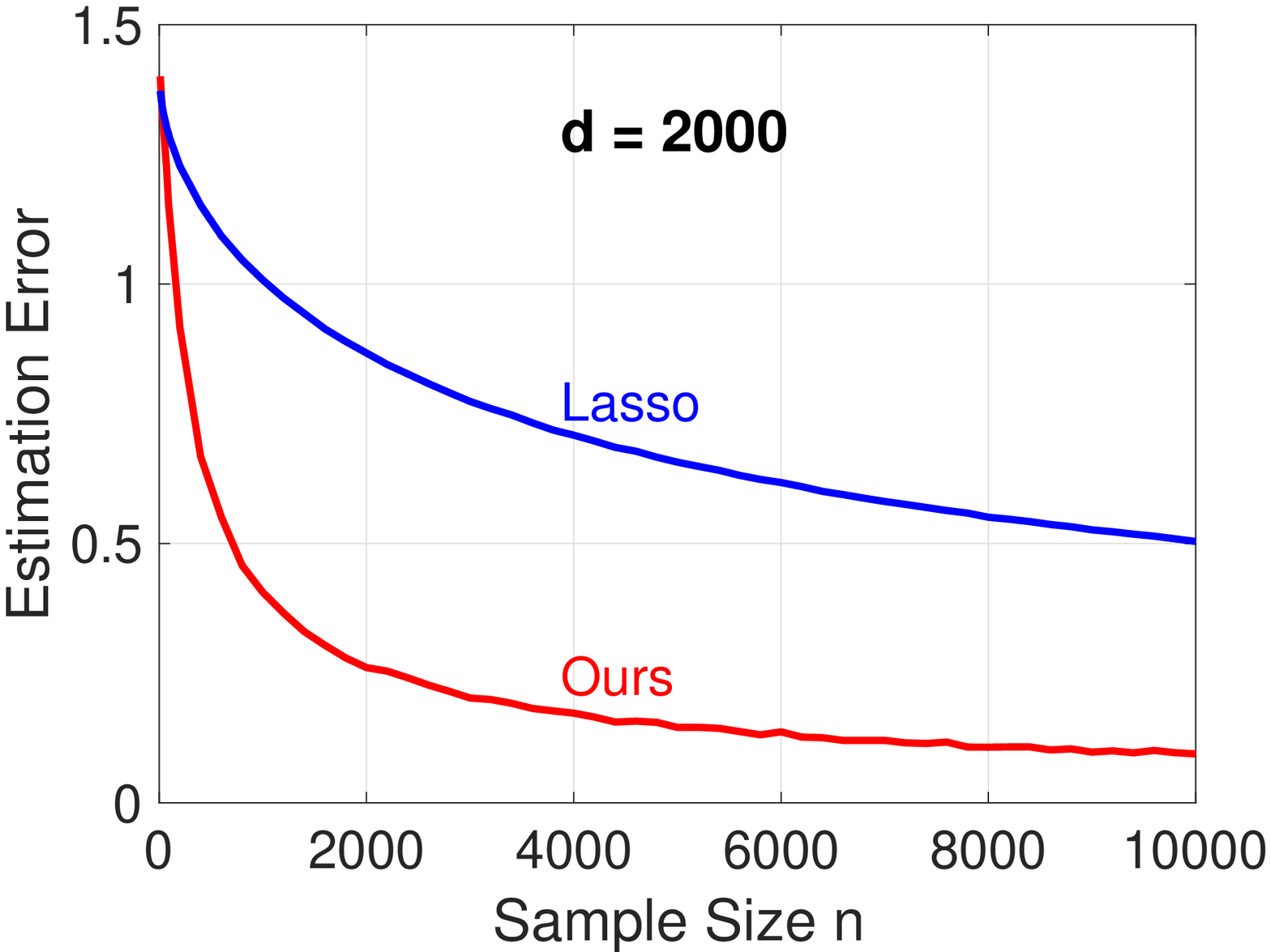}\hspace{0.18in}
\includegraphics[width=0.4\linewidth]{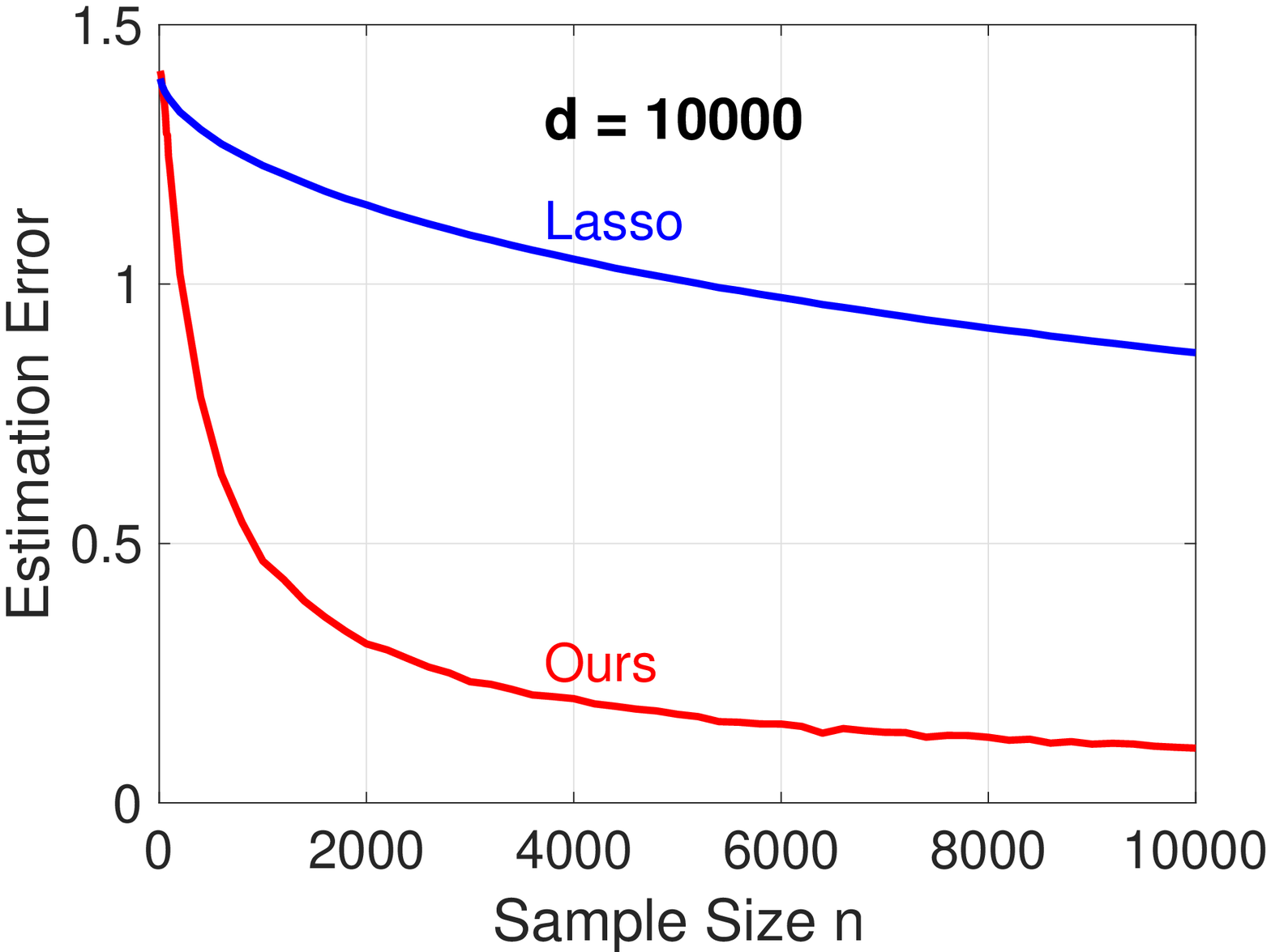}
\vspace{-0.06in}
\caption{{\bfseries Approximation error against sample size.}}
\label{fig:sparse approx}
\end{figure}

\paragraph{Sparse Approximation.}
We study how the estimation error $\twonorm{\xhat - \xbar}$ varies with the sample size $n$. We fix the sparsity $s = 20$, and consider the dimension $d = 2000$ and $d = 10,000$. For each $d$, we increase $n$ from $10$ to $10,000$, and for each configuration of $(d, n)$ we generate the sensing vectors as aforementioned. The error curves are plotted in Figure~\ref{fig:sparse approx}. It shows that the reconstruction error of our method decays much faster than \cite{plan2017high}, and it turns out that their estimator incurs large error when the dimension is increased. In contrast, our algorithm consistently produces accurate estimate.

\begin{figure}[t]
\centering
\includegraphics[width=0.4\linewidth]{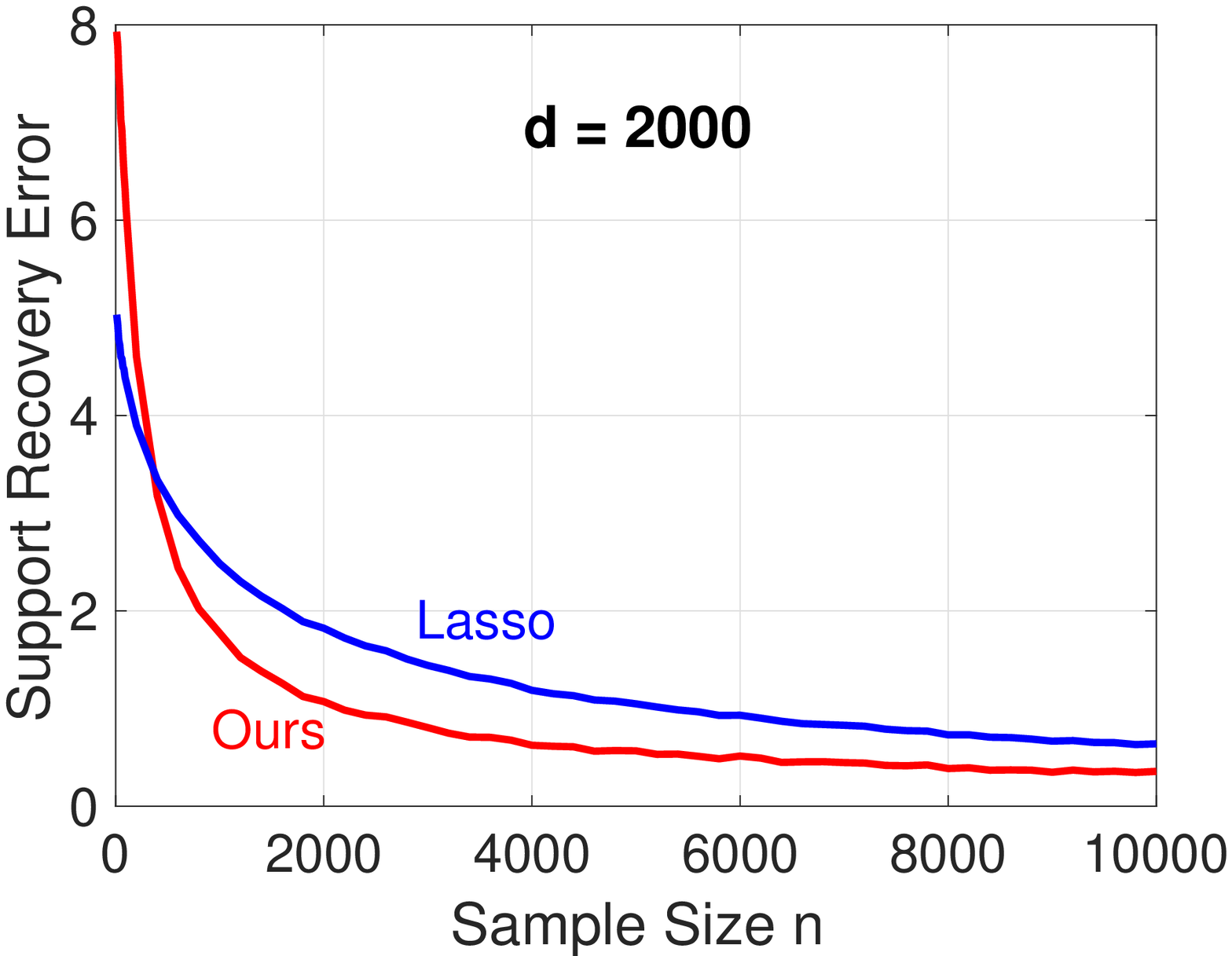}\hspace{0.18in}
\includegraphics[width=0.4\linewidth]{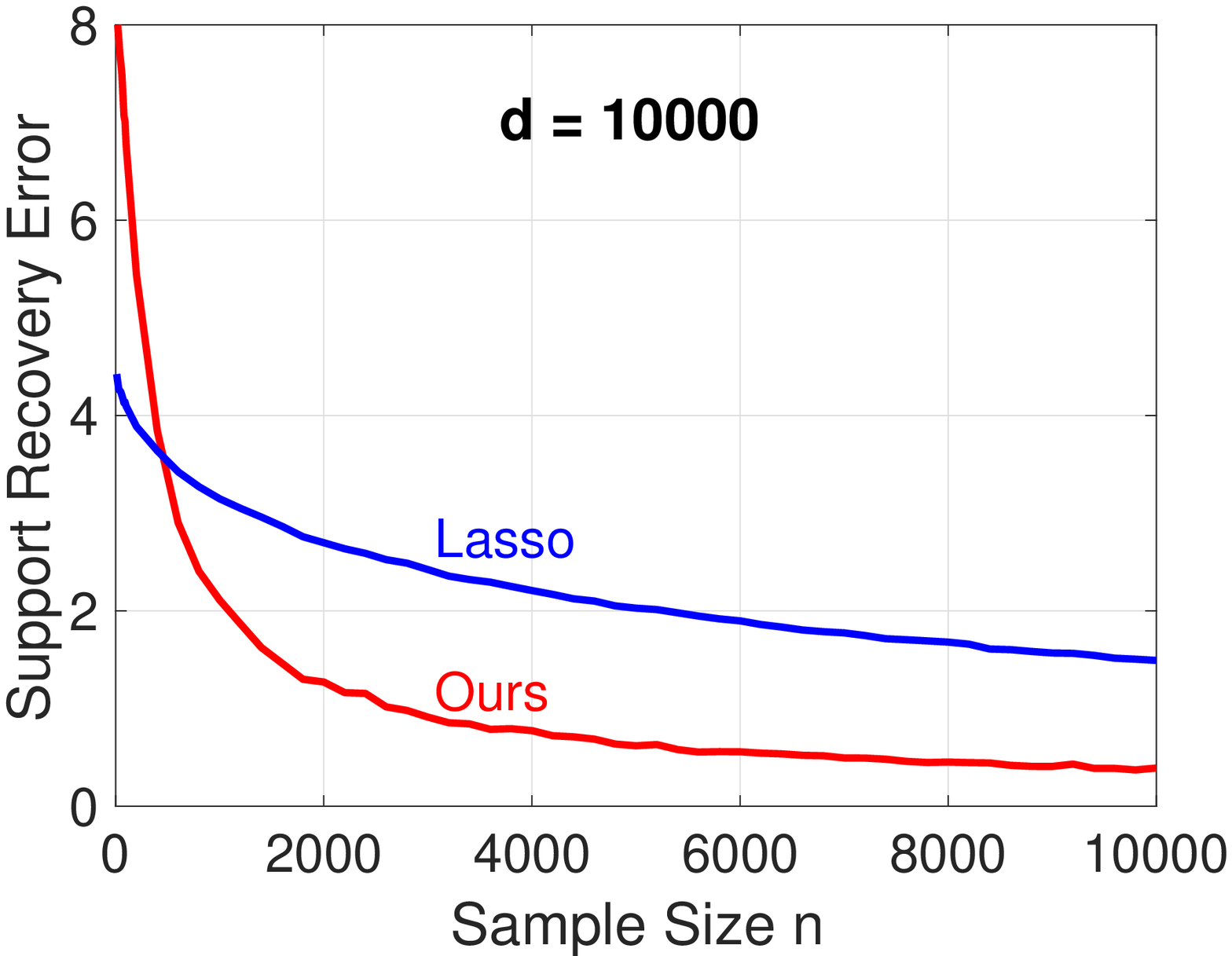}
\vspace{-0.06in}
\caption{{\bfseries Support recovery error against sample size.}}
\label{fig:supp}
\end{figure}

\paragraph{Support Recovery.}
We use the same setting as in sparse approximation, but plot the cardinality of the symmetric difference between $\supp{\xhat}$ and $\supp{\xbar}$. Since \cite{plan2017high} may output arbitrarily dense solution, for fair comparison we apply hard thresholding with the true sparsity to their original estimate. Note that if their original estimate already recovers the support, our post-processing does not hurt it. In this way, the maximum error of both methods is $40$ (since $s = 20$), and zero error indicates perfect support recovery. Again, we observe in Figure~\ref{fig:supp} that our estimator outperforms the state-of-the-art.

\begin{figure}[t]
\centering
\includegraphics[width=0.4\linewidth]{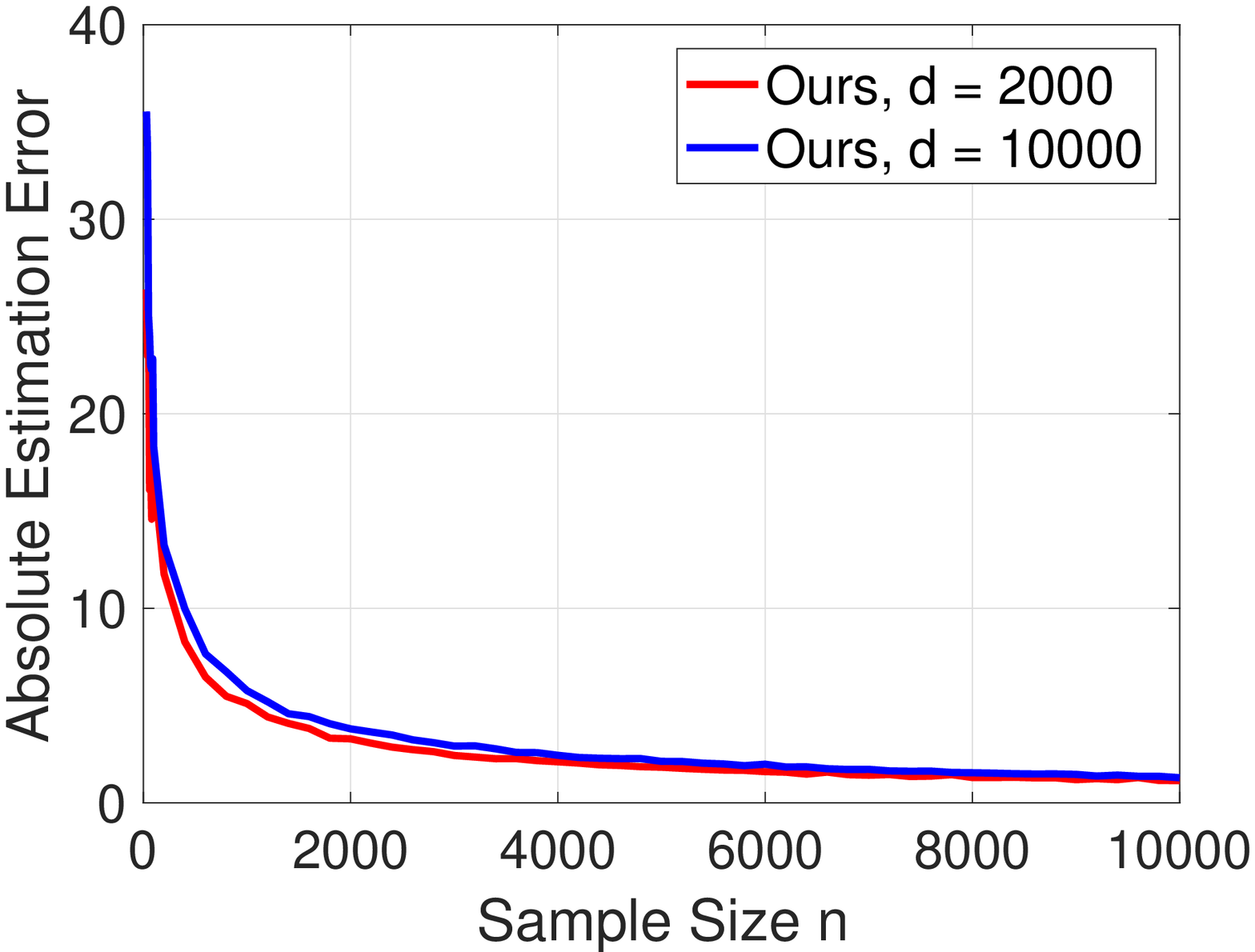}\hspace{0.18in}
\includegraphics[width=0.4\linewidth]{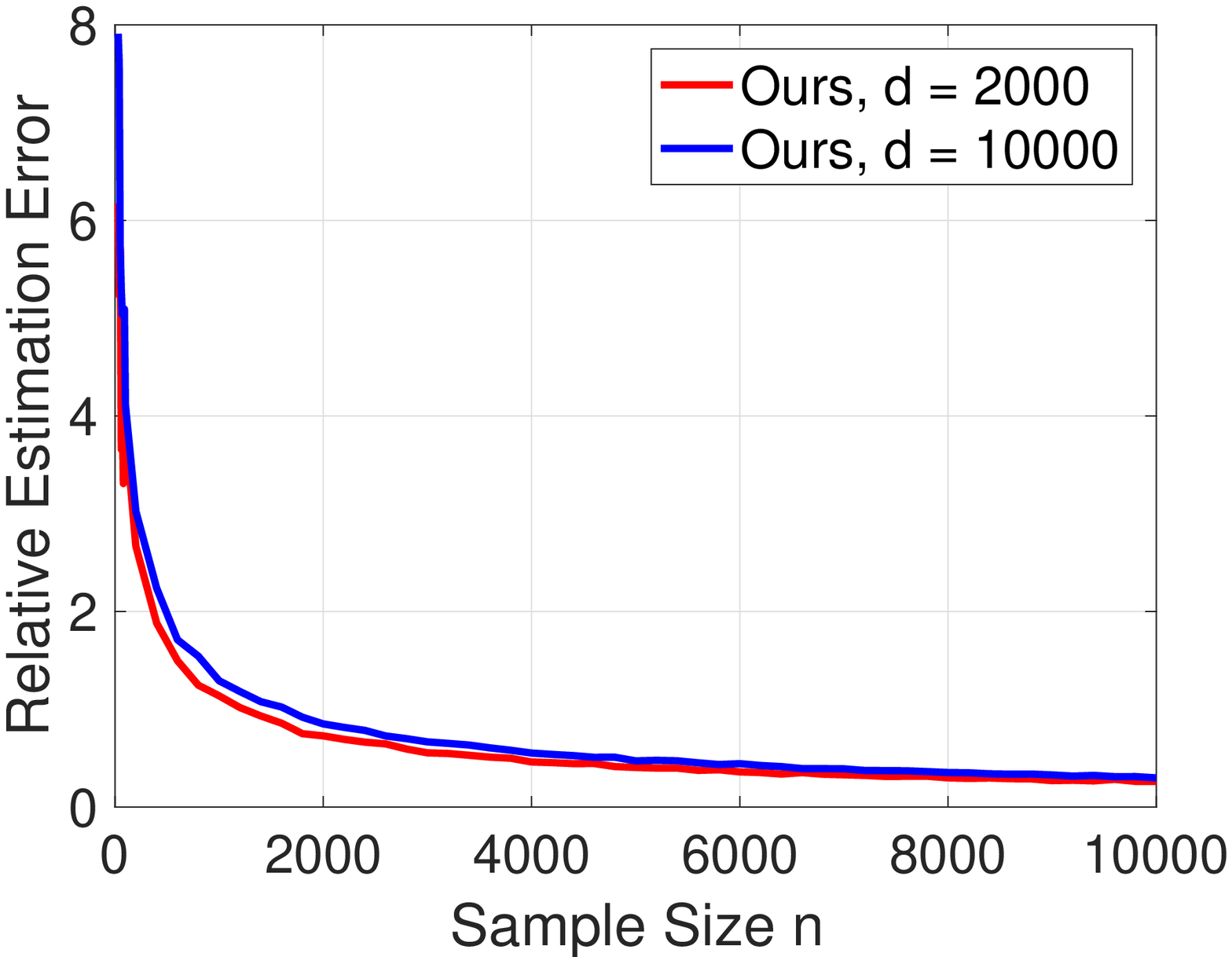}
\vspace{-0.06in}
\caption{{\bfseries Norm estimation error against sample size.}}
\label{fig:norm}
\end{figure}

\paragraph{Norm Estimation.}
We use the estimate presented in Theorem~\ref{thm:norm} to approximate the signal $\xbar$ with $\twonorm{\xbar} \leq R$. We choose $R = 2 \twonorm{\xbar}$, and illustrate the absolute and relative errors in Figure~\ref{fig:norm}. Note that we did not compare with \cite{plan2017high} because norm estimation was not addressed therein. The figure shows that once we have sufficient samples, it is possible to accurately evaluate the norm with the data collection scheme in Section~\ref{subsec:norm}.

\begin{figure}[h]
\centering
\includegraphics[width=0.4\linewidth]{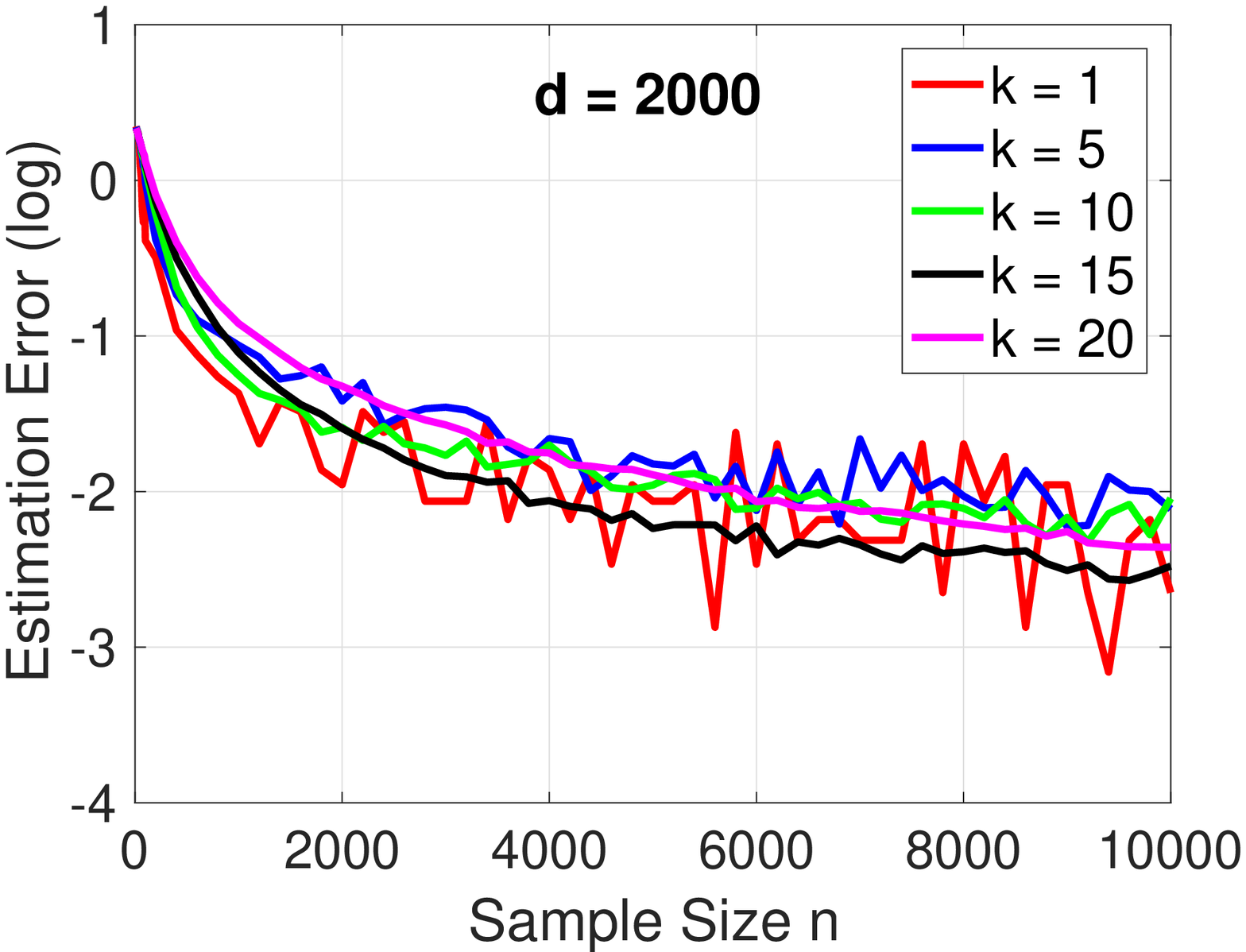}\hspace{0.18in}
\includegraphics[width=0.4\linewidth]{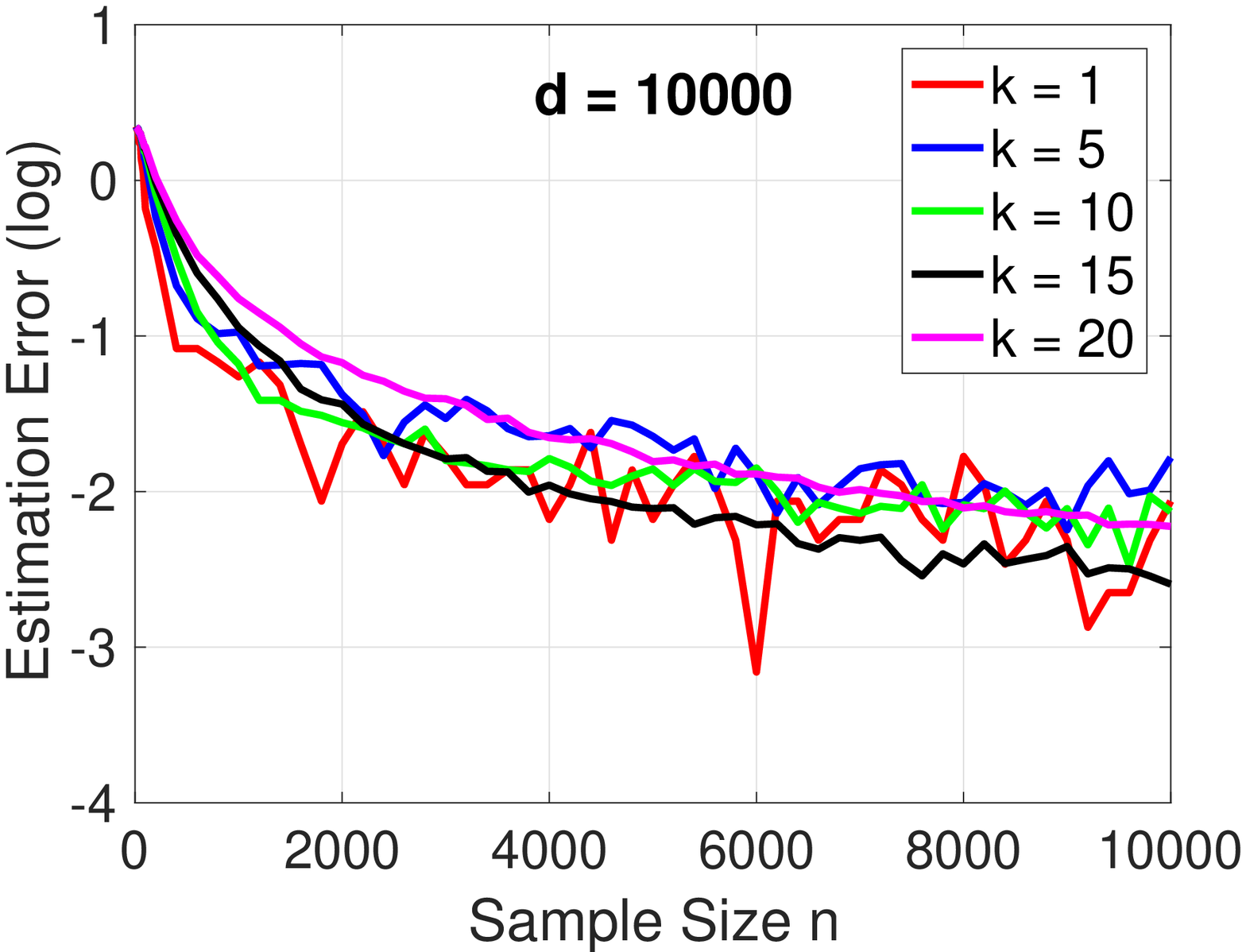}
\vspace{-0.06in}
\caption{{\bfseries Approximation error of the leading $k$ components of our method under model misspecification.} The true sparsity is $20$.}
\label{fig:misspec}
\end{figure}

\paragraph{Model Misspecification.}
In Figure~\ref{fig:misspec} we record the estimation error (in logarithmic scale) of the direction of $\Hk{\xbar}$ when $k < s$. We observe that even in this challenging scenario our estimate possesses a small error. This matches our theoretical guarantee that our estimator is resilient to model misspecification. It is noticeable that the curve of $k = 1$ bumps more often than others. The reason is that when $k=1$, the problem boils down to hitting the unique non-zero position, and the estimation error per signal is either $0$ or $2$ (which results in the bumping phenomenon).

\section{Conclusion}\label{sec:con}
In this paper, we have studied an efficient estimator for recovering a sparse signal from its binary measurements. On the computational side, the estimate can be obtained by a one-step hard thresholding operator which enjoys economic computational and memory cost. On the statistical side, we have shown that the estimation error matches the information-theoretic lower bound up to some logarithmic factor. We have also extended our results to support recovery and norm estimation, and have proved near-optimal error rate in these scenarios. For the estimation of all the three facets of a sparse signal, we have offered rigorous theoretical evidence that our estimator is robust to model misspecification. Finally, we have demonstrated through a comprehensive set of experiments that the practical performance of our estimator matches perfectly our analysis.

\subsubsection*{Acknowledgements}

We thank Jing Wang and Chicheng Zhang for insightful discussions, and thank the anonymous reviewers for helpful comments. This work is supported by the startup funding from Stevens Institute of Technology.


\appendix

\section{Omitted Proofs}\label{sec:app:proof}

This section provides a detailed proof for all the theoretical results.

\subsection{Proof of Proposition~\ref{prop:xhat solution}}\label{sec:app:proof of xhat}
\begin{proof}
Suppose that $\bx$ is a feasible solution to~\eqref{eq:main prog}. Let $\calS = \supp{\bx}$. Consider the objective function value
\begin{align*}
\by\trans\bA \bx \leq \twonorm{\(\bA\trans\by\)_{\calS}} \cdot \twonorm{\bx} = \twonorm{\(\bA\trans\by\)_{\calS}},
\end{align*}
where $\(\bA\trans\by\)_{\calS}$ is interpreted as the $\abs{\calS}$-dimensional vector by truncating the elements of $\bA\trans\by$ outside the index set $\calS$. Evidently, the upper bound is maximized by the choice $\calS = \supp{\bA\trans\by, k}$, and can be attained by picking $\bx = \xhat$ as given in the proposition.
\end{proof}

\subsection{Proof of Lemma~\ref{lem:mean}}\label{sec:app:proof of mean}
The proof follows closely from \cite{zhang2014efficient}.
\begin{proof}
As $\xbar$ is treated as a fixed signal, we have
\begin{align}\label{eq:tmp:1}
\inner{\xbar}{\EXP[y_i \ba_i]} &= \EXP[ y_i \inner{\ba_i}{\xbar} ] \notag\\
&= \EXP\Big[ \EXP \big[ y_i \inner{\ba_i}{\xbar} \mid \ba_i \big] \Big]\notag\\
&= \EXP\Big[ \inner{\ba_i}{\xbar} \EXP \big[ y_i \mid \ba_i \big]  \Big] \notag\\
&= \EXP\big[ \inner{\ba_i}{\xbar} \theta\( \inner{\ba_i}{\xbar} \) \big] \notag\\
&= \EXP_{g \sim N(0, 1)}\big[ g \cdot \theta_i(g) \big] = \lambda_i.
\end{align}
In the above expression, the second equality is by law\ of\ total\ expectation, the third equality is by Assumption~\ref{as:indep}, the fourth equality is by Model~\eqref{eq:model yi}, and the last equality applies~\ref{as:ai} and $\twonorm{\xbar}=1$.

On the other side, by Gram-Schmidt process we can obtain an orthonormal basis $\{\xbar, \bv_1, \dots, \bv_{d-1}\}$. For any $\bv_j$, it follows that $\inner{\ba_i}{\bv_j}$ is independent from $\inner{\ba_i}{\xbar}$, which implies for all $1 \leq j \leq d-1$
\begin{equation}\label{eq:tmp:2}
\inner{\bv_j}{\EXP[y_i \ba_i]} = \EXP[y_i \inner{\ba_i}{\bv_j}] = \EXP[y_i] \cdot \EXP[\inner{\ba_i}{\bv_j}] = 0 .
\end{equation}
Now for the sake of contradiction suppose that $\EXP[y_i \ba_i] = \lambda_i \xbar + \bu$ with $\bu \neq \bzero$. After plugging this expression of $\EXP[y_i \ba_i]$ into \eqref{eq:tmp:1} and \eqref{eq:tmp:2}, we have that $\bu$ is orthogonal to the orthonormal basis, which implies that $\bu$ must be a zero vector. This yields a contradiction.
\end{proof}

\subsection{Proof of Lemma~\ref{lem:inf norm}}
We use well-known concentration inequalities of Gaussian random variables to prove the result.
\begin{proof}
Recall that $\bA = (\ba_1, \dots, \ba_n)\trans$ and $\by = (y_1, \dots, y_n)\trans$. Thus
\begin{equation*}
\frac{1}{n} \bA\trans \by = \frac{1}{n} \sum_{i=1}^{n} \ba_i y_i.
\end{equation*}
Consider the coordinate index $j \in \{1, \dots, d\}$. The $j$th coordinate of $\frac{1}{n} \bA\trans \by - \lambda \xbar$ is given by
\begin{equation*}
\(\frac{1}{n} \bA\trans \by - \lambda \xbar\)^{(j)} = \frac{1}{n} \sum_{i=1}^n \Big[ a_i^{(j)} y_i - \lambda_i \bar{x}^{(j)} \Big]
\end{equation*}
where $a_i^{(j)}$ and $\bar{x}^{(j)}$ are the $j$th coordinate of $\ba_i$ and $\xbar$ respectively. For a random variable $v$, define the norm
\begin{equation*}
\lVert v \rVert_{\psi_2} \defeq \sup_{p \geq 1} p^{-1/2} \( \EXP \abs{v}^p \)^{1/p}.
\end{equation*}
It is not hard to see that $\lVert a_i^{(j)} y_i \rVert_{\psi_2} = \lVert a_i^{(j)} \rVert_{\psi_2} \leq \const_0$ where the inequality holds since $a_i^{(j)}$ is standard normal~\cite{vershynin2010introduction}. Furthermore, for any random variable $v$ we have
\begin{align*}
\lVert v - \EXP v \rVert_{\psi_2} \leq \lVert v\rVert_{\psi_2} + \lVert \EXP v\rVert_{\psi_2} = \lVert v \rVert_{\psi_2} + \abs{\EXP v}\leq \lVert v \rVert_{\psi_2} + \EXP\abs{v} \leq 2 \lVert v \rVert_{\psi_2}.
\end{align*}
Thus $\lVert a_i^{(j)} y_i - \lambda_i \bar{x}^{(j)} \rVert_{\psi_2} \leq 2\const_0$. Now by Proposition~5.10 in \cite{vershynin2010introduction} we have
\begin{align*}
\Pr\( \abs{ \frac{1}{n} \sum_{i=1}^{n} \Big[ a_i^{(j)} y_i - \lambda_i \bar{x}^{(j)} \Big]  } \geq t \) \leq e \cdot \exp\( - \frac{\const_1 t^2 }{(2 \const_0)^2 / n} \).
\end{align*}
Picking $t = \const \sqrt{ \frac{\log d}{n}}$ for some constant $\const$ we have
\begin{align*}
\Pr\( \abs{ \frac{1}{n} \sum_{i=1}^{n} \Big[ a_i^{(j)} y_i - \lambda_i \bar{x}^{(j)} \Big]  } \geq  \const \sqrt{ \frac{\log d}{n}}\) \leq \exp\(- 11 \log d\) = d^{-11}.
\end{align*}
By union bound over the coordinates $j$, we show that
\begin{equation*}
\Pr\( \infnorm{\frac{1}{n} \bA\trans \by - \lambda \xbar } \geq \const \sqrt{\frac{\log d}{n}}  \) \leq d^{-10}
\end{equation*}
which completes the proof.
\end{proof}

\subsection{Proof of Theorem~\ref{thm:main} and Theorem~\ref{thm:sparse approx misspec}}\label{sec:app:proof of approx}
The proof is very modular and thus facilitates the extension to other types of signals and sensing matrices, as we discussed in Remark~\ref{rmk:decomp}.
\begin{proof}
Consider the program~\eqref{eq:main prog}. Note that $\xhat$ is optimal and $\Hk{\xbar} / \twonorm{\Hk{\xbar} }$ is feasible. It is worth mentioning that when $k \geq s$, $\Hk{\xbar} / \twonorm{\Hk{\xbar} } = \xbar$. For simplicity, we write $\bz = { \Hk{\xbar}}/{\twonorm{\Hk{\xbar}}}$. It immediately follows that
\begin{equation*}
\by\trans \bA \xhat \geq \by\trans \bA \bz.
\end{equation*}
With some re-arrangement, we have
\begin{equation*}
 \inner{- \lambda \bz}{\xhat - \bz} \leq \inner{\frac{1}{n} \bA\trans \by - \lambda \bz}{\xhat - \bz}.
\end{equation*}
Note that as $\twonorm{\bz} = \twonorm{\xhat} = 1$, the left-hand side can be written as follows:
\begin{equation*}
\inner{-\lambda \bz}{\xhat - \bz} = \frac{\lambda}{2} \twonorm{\xhat - \bz}^2.
\end{equation*}
Thus, we obtain
\begin{align*}
\frac{\lambda}{2} \twonorm{\xhat - \bz}^2 \leq \inner{\frac{1}{n} \bA\trans \by - \lambda \bz}{\xhat - \bz} &\leq \infnorm{\frac{1}{n} \bA\trans \by - \lambda \bz } \cdot \onenorm{ \xhat - \bz} \\
&\leq \infnorm{\frac{1}{n} \bA\trans \by - \lambda \bz } \cdot \sqrt{2k} \twonorm{\xhat - \bz},
\end{align*}
where the second inequality follows from H\"older's inequality, the third inequality follows from the facts that $\onenorm{\bv} \leq \sqrt{\zeronorm{\bv}} \cdot \twonorm{\bv}$ for all $\bv \in \Rd$ and that $\zeronorm{\xhat - \bz} \leq 2k$. Dividing both sides by $\twonorm{\xhat - \bz}$ gives
\begin{align*}
\twonorm{\xhat - \bz} \leq \frac{\sqrt{2k}}{\lambda} \infnorm{\frac{1}{n} \bA\trans \by - \lambda \bz } \leq { \frac{\sqrt{2k}}{\lambda} \infnorm{\frac{1}{n} \bA\trans \by - \lambda \xbar } } + { \sqrt{2k} \infnorm{\bz - \xbar} }.
\end{align*}
The first term on the right-hand side can be bounded from the above by Lemma~\ref{lem:inf norm}. 
\end{proof}

\subsection{Proof of Lemma~\ref{lem:supp recovery}}\label{sec:app:proof of supp}
This is a well recognized result, and we include the proof for completeness.

\begin{proof}
For the sake of contradiction, suppose that $\supp{\xbar}$ is not contained in $\supp{\bx}$. That is, there exists an index $j \in \{ 1, \dots, d \}$, such that $\bar{x}_j \neq 0, \quad x_j = 0$. Therefore, we have $\twonorm{\bx - \xbar} \geq \abs{\bar{x}_j} \geq \xbar_{\min}$, which leads to a contradiction.
\end{proof}

%
%

\subsection{Proof of Theorem~\ref{thm:norm}}\label{sec:app:proof of norm}

\begin{proof}
Consider the program~\eqref{eq:aug prob}. Let the global optimum $\xhat' = \frac{(\bx_0;\ t_0)}{\sqrt{\twonorm{\bx_0}^2 + t_0^2}}$ where $\bx_0 \in \Rd$ and $t_0 \in \R$. Since $\xbar'$ is feasible, we may apply Theorem~\ref{thm:main} and obtain that with probability at least $1 - d^{-10}$,
\begin{equation}\label{eq:tmp}
\twonorm{\frac{(\xbar;\ R)}{\sqrt{\twonorm{\xbar}^2 + R^2}} - \frac{(\bx_0;\ t_0)}{\sqrt{\twonorm{\bx_0}^2 + t_0^2}}} \leq \frac{\const}{\lambda} \sqrt{\frac{k \log d}{n}} =: {\delta}.
\end{equation}
For the case $t_0 \neq 0$, the proof of Theorem~4 in \cite{knudson2016bit} immediately implies that
\begin{equation*}
\twonorm{\frac{R}{t_0} \bx_0 - \xbar} \leq \frac{\const_1 \cdot R}{\lambda} \sqrt{\frac{k \log d}{n}}.
\end{equation*}
For the case $t_0 = 0$, it follows from \eqref{eq:tmp}
\begin{equation}
\twonorm{ \frac{ (\bx_1;\ 1) }{\sqrt{ \twonorm{\bx_1}^2 + 1 }} - { (\bx_2;\ 0) } } \leq \delta,
\end{equation}
where we write $\bx_1 = \xbar / R$ and $\bx_2 = \bx_0 / \twonorm{\bx_0}$.

On the other side, we have
\begin{align*}
\twonorm{\bx_1 - \bx_2} &= \sqrt{\twonorm{\bx_1}^2 + 1} \twonorm{ \frac{\bx_1}{ \sqrt{\twonorm{\bx_1}^2 + 1} } - \frac{\bx_2}{ \sqrt{\twonorm{\bx_1}^2 + 1} }  }\\
&\stackrel{\zeta_1}{\leq} \sqrt{2} \twonorm{ \frac{\bx_1}{ \sqrt{\twonorm{\bx_1}^2 + 1} } - \frac{\bx_2}{ \sqrt{\twonorm{\bx_1}^2 + 1} }  }\\
&\stackrel{\zeta_2}{\leq} \sqrt{2} \twonorm{ \frac{\bx_1}{ \sqrt{\twonorm{\bx_1}^2 + 1} } - \bx_2  } + \sqrt{2} \twonorm{ \bx_2 - \frac{\bx_2}{ \sqrt{\twonorm{\bx_1}^2 + 1} }  }\\
&\stackrel{\zeta_3}{\leq} \sqrt{2} \twonorm{ \frac{\bx_1}{ \sqrt{\twonorm{\bx_1}^2 + 1} } - \bx_2 } + \sqrt{2} \abs{ 1 - \frac{1}{ \sqrt{\twonorm{\bx_1}^2 + 1} }  }\\
&\stackrel{\zeta_4}{\leq} \sqrt{2} \twonorm{ \frac{\bx_1}{ \sqrt{\twonorm{\bx_1}^2 + 1} } - \bx_2 } + \sqrt{2} \abs{  \frac{1}{ \sqrt{\twonorm{\bx_1}^2 + 1} }  }\\
&\stackrel{\zeta_5}{\leq} 2 \(  \twonorm{ \frac{\bx_1}{ \sqrt{\twonorm{\bx_1}^2 + 1} } - \bx_2  }^2 +  \abs{  \frac{1}{ \sqrt{\twonorm{\bx_1}^2 + 1} }  }^2 \)^{1/2}\\
&= 2 \twonorm{ \frac{ (\bx_1;\ 1) }{\sqrt{ \twonorm{\bx_1}^2 + 1 }} - (\bx_2;\ 0) }\\
&\leq 2 \delta.
\end{align*}
In the above, the inequality $\zeta_1$ holds since $\twonorm{\bx_1} \leq 1$, $\zeta_2$ follows from the triangle inequality, $\zeta_3$ uses the fact $\twonorm{\bx_2} \leq 1$. To see why $\zeta_4$ holds, note that $\twonorm{\bx_1} \leq 1$ implies
\begin{equation*}
\abs{1 - \frac{1}{ \sqrt{\twonorm{\bx_1}^2 + 1} } } \leq 1 - 1/\sqrt{2} < 1 / \sqrt{2} \leq \frac{1}{ \sqrt{\twonorm{\bx_1}^2 + 1}}.
\end{equation*}
The last inequality $\zeta_5$ applies $a + b \leq \sqrt{2} \sqrt{a^2 + b^2}$ for any two scalars $a$ and $b$. Therefore, we obtain that
\begin{equation*}
\twonorm{\frac{\xbar}{R} - \frac{\bx_0}{\twonorm{\bx_0}}} \leq \frac{2\const}{\lambda} \sqrt{\frac{k\log d}{n}},
\end{equation*}
namely
\begin{equation*}
\twonorm{ \frac{R}{\twonorm{\bx_0}} \bx_0 - \xbar } \leq \frac{2\const \cdot R}{\lambda} \sqrt{\frac{k\log d}{n}}
\end{equation*}
which completes the proof.
\end{proof}